\documentclass[10pt,twocolumn,letterpaper]{article}

\usepackage{cvpr}
\usepackage{times}
\usepackage{epsfig}
\usepackage{graphicx}

\usepackage{amsmath}
\usepackage{amssymb}

\usepackage{bm}
\usepackage{dsfont}

\usepackage[pagebackref=true,breaklinks=true,letterpaper=true,colorlinks,bookmarks=false]{hyperref}

\cvprfinalcopy 

\def\cvprPaperID{2168} 

\ifcvprfinal\fi
\begin{document}
\title{ABC-CNN: An Attention Based Convolutional Neural Network for Visual Question Answering}

\author{Kan Chen\\
University of Southern California\\
{\tt\small kanchen@usc.edu}
\and
Jiang Wang\\
Baidu Research - IDL\\
{\tt\small wangjiang03@baidu.com}
\and
Liang-Chieh Chen\\
UCLA\\
{\tt\small lcchen@cs.ucla.edu}
\and
Haoyuan Gao\\
Baidu Research - IDL\\
{\tt\small gaohaoyuan@baidu.com}
\and
Wei Xu\\
Baidu Research - IDL\\
{\tt\small wei.xu@baidu.com}
\and
Ram Nevatia\\
University of Southern California\\
{\tt\small nevatia@usc.edu}
}

\maketitle

\begin{abstract}
We propose a novel attention based deep learning architecture for visual question answering task (VQA).
Given an image and an image-related question, VQA returns a natural language answer.
Since different questions inquire about the attributes of different image regions, generating correct answers requires the model to have question-guided attention,
i.e., the attention on the regions corresponding to the input question's intent.
We introduce an attention-based configurable convolutional neural network (ABC-CNN) to locate the question-guided attention based on input queries.
ABC-CNN determines the attention regions by finding the corresponding visual features in the visual feature maps with a ``configurable convolution'' operation.
With the help of the question-guided attention, ABC-CNN can achieve both higher VQA accuracy and better understanding of the visual question answering process.
We evaluate the ABC-CNN architecture on three benchmark VQA datasets: Toronto COCO-QA, DAQUAR, and VQA dataset. ABC-CNN model achieves significant improvements over state-of-the-art methods.
The question-guided attention generated by ABC-CNN is also shown to be the regions that are highly relevant to the questions' intents.
\end{abstract}

\section{Introduction}
Visual Question Answering (VQA) is the task of answering questions, posed in natural language, about the semantic content in an image (or video).
Given an image and an image related question, VQA answers the question in one word or a natural language sentence.
VQA is of great importance to many applications, including image retrieval, early education, and navigation for blind people as it provides user-specific information through the understanding of both the natural language questions and image content.
VQA is a highly challenging problem as it requires the machine to understand natural language queries, extract semantic contents from images, and relate them in a unified framework.
In spite of these challenges, an exciting set of methods have been developed by the research community in recent years.

\begin{figure}
\begin{center}
\includegraphics[width=3.3in]{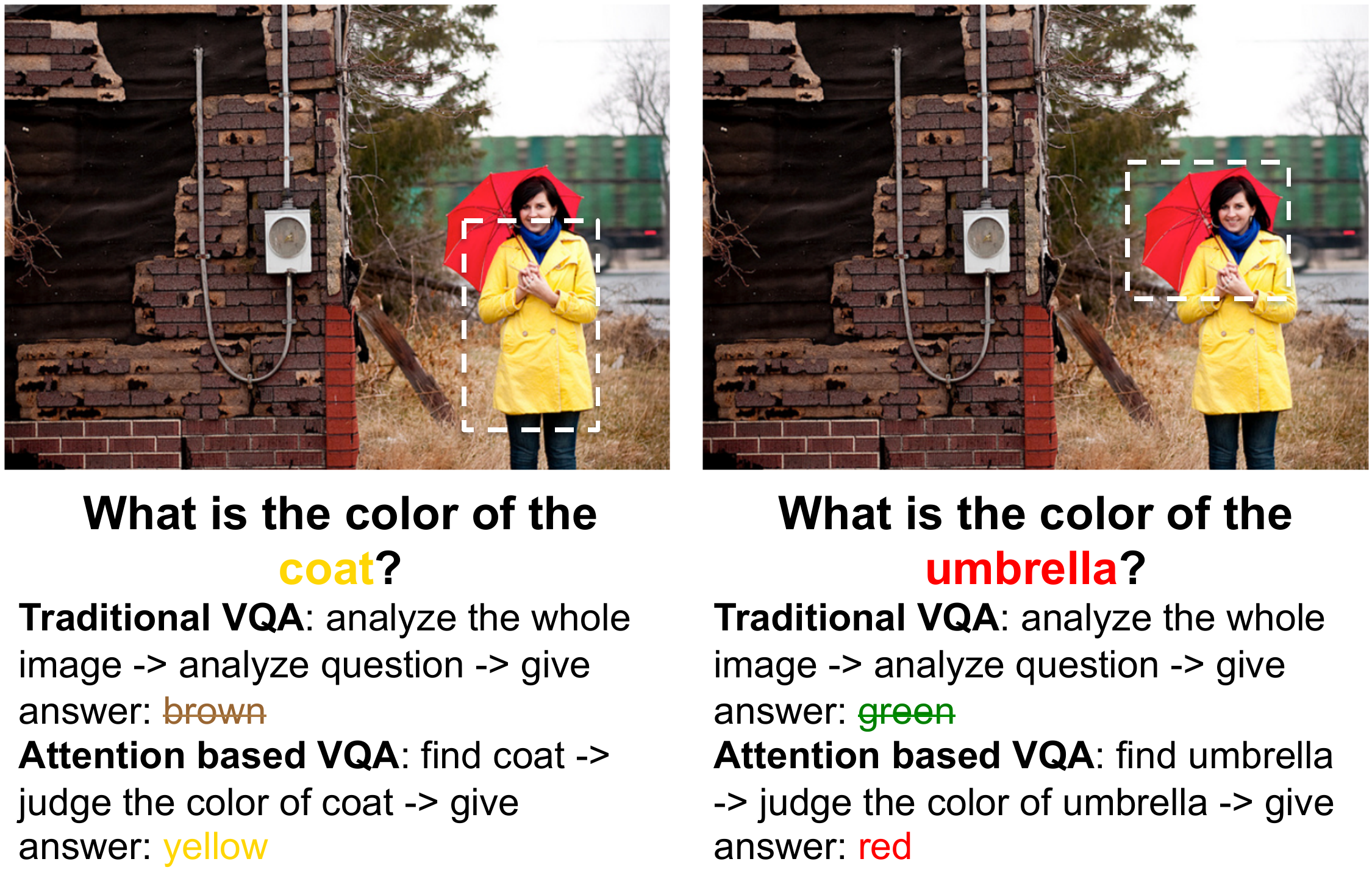}
\end{center}
   \caption{Attention in visual question answering. For different questions, the corresponding attention region varies from white dashed box ``coat'' in the left image to the one ``umbrella'' in the right image.}
\label{fig: example of attention}
\end{figure}

Current state-of-the-art VQA models \cite{torontoQa}\cite{malinowski2015ask}\cite{gao2015you} generally contain a vision part, a question understanding part and an answer generation part.
The vision part extracts visual features through a deep convolutional neural network (CNN) \cite{lecun1989backpropagation} or using a traditional visual feature extractor.
The question understanding part learns a dense question embedding feature vector to encode question semantics, either with a Bag-of-Words model~\cite{torontoQa} or a recurrent neural network (RNN) ~\cite{hochreiter1997long} model.
The answer generation part produces an answer conditioned on the visual features and the question embeddings.
The answer can either be a single word generated by a multi-class classifier \cite{torontoQa} or a full sentence generated by an additional RNN decoder  \cite{malinowski2015ask}\cite{gao2015you}.
The visual features and dense question embeddings are integrated through a linear~\cite{torontoQa} / non-linear~\cite{malinowski2015ask}\cite{gao2015you} transform which jointly projects the features from image space and semantic space into answer space.
This integration is normally not sufficient to fully exploit the relationship of the vision part and the question understanding part
because it loses the opportunity to exploit the intent of queries to focus on different regions in an image.

When trying to answer a question about an image, humans tend to search the informative regions according to the question's intent before giving the answer.
For example, in Fig.~\ref{fig: example of attention}, considering the query ``What is the color of the coat?'', it is common for humans to focus attention on the region of coat before judging its color to answer the question.
Based on this observation, we propose a novel attention-based configurable convolutional neural network (ABC-CNN) to locate such informative regions and give more correct answers for VQA.
We call the mechanism of finding informative regions based on the input question's intent as ``\emph{question-guided attention}'', because these regions are determined by both images and image-related questions.

\begin{figure*}
\begin{center}
\includegraphics[width=7in]{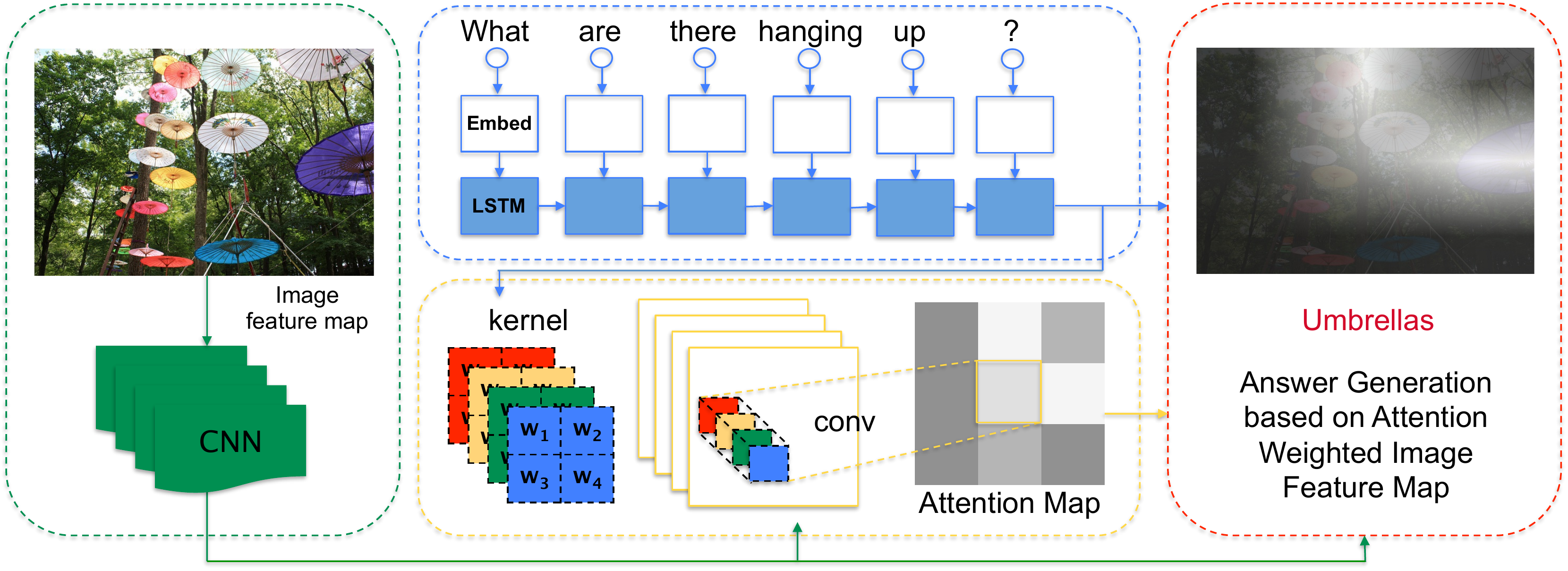}
\end{center}
   \caption{The framework of ABC-CNN. The green box denotes the image feature extraction part using CNN; the blue box is the question understanding part using LSTM; the yellow box illustrates the attention extraction part with configurable convolution; the red box is the answer generation part using multi-class classification based on attention weighted image feature maps. The orange letters are corresponding variables explained in Eq. (\ref{equ: kernel learn}) - (\ref{equ: ans generate}).}
\label{fig: framework}
\end{figure*}

As shown in Fig.~\ref{fig: framework}, ABC-CNN contains a vision part, a question understanding part, an answer generation part, and an attention extraction part.
We employ a CNN to extract visual features in the vision part. Instead of extracting a single global visual feature, we extract a spatial feature map to retain crucial spatial information, by either applying a CNN in a sliding window way or with a fully convolutional neural network.
A long-short term memory (LSTM)~\cite{hochreiter1997long} model is used to obtain question embeddings in the question understanding part.
In this paper, we only consider the VQA task with single word answers which can be generated by utilizing a multi-class classifier in the answer generating part.
Our method can be extended to generate full sentences by using an RNN decoder.

We present the question-guided attention information as a question-guided attention map (QAM), which is the core of the ABC-CNN framework.
We model the QAM as latent information, and do not require explicit labeling of such maps for all kinds of possible queries.
The QAM is generated by searching for visual features that correspond to the input query's semantics in the spatial image feature map.
We achieve the search via a configurable convolution neural network, which convolves the visual feature map with a configurable convolutional kernel (CCK).
This kernel is generated by transforming the question embeddings from the semantic space into the visual space, which contains the visual information determined by the intent of the question.
For example, in Fig.~\ref{fig: example of attention}, the question ``what is the color of the umbrella?'' should generate a CCK that corresponds to the ``umbrella'' visual features.
Convolving the CCK with image feature map adaptively represents each region's importance for answering the given question as a QAM.
The generated QAMs can be utilized to spatially weight the visual feature maps to filter out noise and unrelated information.
With the visual features conditioned on the input query, ABC-CNN can return more accurate answers from the multi-class classifier in answer generation part.
The whole framework can be trained in an end-to-end way without requiring any manual labeling of attention regions in images.

In the experiments, we evaluate the ABC-CNN framework on three benchmark VQA datasets: Toronto COCO-QA \cite{torontoQa}, DAQUAR  \cite{malinowski2014nips} and VQA~\cite{antol2015vqa}.
Our method significantly outperforms state-of-the-art methods.
Visualization of attention maps demonstrates that the ABC-CNN architecture is capable of generating attention maps that well reflect the regions queried by questions.

In summary, we propose a unified ABC-CNN framework to effectively integrate the visual and semantic information for VQA via question-guided attention.
Not only does the question guided attention significantly improve the performance of VQA systems, but it also helps us gain a better understanding of the question answering process.

\section{Related Work}
\textbf{Image captioning:} VQA and image captioning are highly related because both of them need to reason about the visual contents and present the results in a full natural language sentence or in a word.
Current state-of-the-art methods in VQA \cite{gao2015you}\cite{malinowski2015ask}\cite{torontoQa} and image captioning \cite{mao2014deep}\cite{donahue2014long}\cite{karpathy2014deep}\cite{venugopalan2014translating}\cite{xu2015show}\cite{zitnick2013learning} generally apply a CNN to extract visual features and an LSTM model as a decoder to generate answers or captions.
\cite{gao2015you}\cite{mao2014deep}\cite{malinowski2015ask} apply a multi-modal layer to combine the visual features and word embedding vectors by a joint projection during the caption generation in the LSTM decoder.
\cite{torontoQa} employs the projected image features as the starting states of the LSTM decoder, similar to the encoder-decoder framework in sequence to sequence learning~\cite{sutskever2014sequence}.
Treating image features as global visual features, these studies in VQA and image captioning fail to exploit the valuable information in questions to focus their attention on the corresponding regions in images.

\textbf{Attention models:} Attention models have been successfully adopted in many computer vision tasks, including object detection \cite{mnih2014recurrent}\cite{ba2014multiple}, fine-grained image classification \cite{sermanet2014attention}\cite{lin2015bilinear}, and image captioning \cite{jin2015aligning}.
Attention can be modeled as a region sequence in an image. An RNN model is utilized to predict the next attention region based on the current attention region's location and visual features.
\cite{mnih2014recurrent}, \cite{ba2014multiple} and \cite{sermanet2014attention} employ this framework for object recognition, object detection, and fine-grained object recognition, respectively.
\cite{jin2015aligning} develops an attention-based model for image captioning that uses an RNN as a decoder.
It extracts a set of proposal regions in each image, and learns their attention weights using the decoding LSTM decoder's hidden states and the visual features extracted in each proposal region.
In \cite{lin2015bilinear}, a bilinear CNN structure utilizes two separate branches to combine the location and content information for fine-grained image classification.
ABC-CNN is inspired by the successful application of attention on these vision tasks and utilize  {\em question-guided} attention to improve VQA  performance.

\textbf{Configurable convolutional neural network:}
In \cite{klein2015dynamic}, a dynamic convolutional layer architecture is proposed for short range weather prediction.
The convolutional kernels in the dynamic convolutional layer are determined by a neural network encoding the information of weather images in previous time steps.
In VQA, the most important clue to determine the attention regions is the question.
Thus, the CCKs in ABC-CNN framework are determined by the question embedding.

\section{Attention Based Configurable CNN}
The framework of ABC-CNN is illustrated in Fig.~\ref{fig: framework}.
We restrict to QA pairs with single-word answers in this paper; this allows us to treat the task as a multi-class classification problem, which simplifies the evaluation metrics
so that we can concentrate on developing question-guided attention models.

ABC-CNN is composed of four components: (1) the image feature extraction part, (2) the question understanding part, (3) the attention extraction part and (4) the answer generation part.
In the image feature extraction part (green box), a deep CNN is used to extract an image feature map $\mathbf{I}$ for each image as the image representation.
We utilize the VGG-19 deep convolutional neural network~\cite{simonyan2014very} pretrained on 1000-class ImageNet classification challenge 2012 dataset~\cite{deng2009imagenet}, and a fully convolutional segmentation neural network~\cite{chen2014semantic} pretrained on PASCAL 2007 segmentation dataset.
The question understanding part (blue box) adopts an LSTM to learn a dense question embedding vector $\bm{s}$ to encode semantic information of an image-related question.
The attention extraction part (yellow box) configures a set of configurable convolutional kernels (CCK) according to different dense question embeddings.
These kernels, emphasizing the visual features of objects asked in the question, are convolved with the image feature maps to generate question-guided attention maps (QAM).
The answer generation part, shown in the red box, answers a question using a multi-class classifier based on the image feature map $\mathbf{I}$, the attention weighted image feature map, and the dense question embedding vector $\bm{s}$.
The rest of this section will describe each component of ABC-CNN framework in details.


\subsection{Attention Extraction}\label{subsec: att_learn}
A QAM, $\bm{m}$, capturing the image regions queried by the question, is generated for each image-question pair using a configurable convolutional neural network.
The configurable convolution operation can be thought of as searching spatial image feature maps for specific visual features that correspond to the question's intent.
The specific visual features are encoded as a CCK $\bm{k}$ in this network, which is configured by projecting the dense question embedding $\bm{s}$ from semantic space to visual space.

\begin{equation}\label{equ: kernel learn}
\boldsymbol{k} = \sigma(\mathbf{W}_{sk}\bm{s}+\bm{b}_k),\text{\ \ \ } \sigma(x) = \frac{1}{1+e^{-x}}
\end{equation}
where $\sigma(.)$ is a sigmoid function.

The dense question embedding $\bm{s}$ encodes the semantic object information asked in the question.
The projection transforms the semantic information into the corresponding visual information as a CCK, which has the same number of channels as the image feature map $\mathbf{I}$.

The QAM is generated by convolving the CCK $\bm k$ with the image feature map $\mathbf{I}$,
and applying softmax normalization:

\begin{equation}\label{equ: att map gen}
\bm{m}_{ij} = P(\text{ATT}_{ij}|\mathbf{I}, \bm{s}) = \frac{e^{\bm{z}_{ij}}}{\sum_i\sum_je^{\bm{z}_{ij}}},\text{\ \ \ }\bm{z} = \bm{k}*\mathbf{I}
\end{equation}
where $\bm{m}_{ij}$ is the element of the QAM at position $(i, j)$, and the symbol *  represents the convolution operation.
The QAM characterizes the attention distribution across the image feature map.
The convolution is padded so that the QAM $\bm{m}$ has the same size as the image feature map $\mathbf{I}$.
The QAM corresponds to the regions asked by the question.
For example, the question ``What is the color of the umbrella?'' can generate an attention map focusing on umbrella image regions
because the CCK is configured to find umbrella visual features.

%
With the attention map $\bm{m}$, we can improve the question answering accuracy on  various classes of questions for the following reasons:
\begin{itemize}
\item For counting questions, such as ``how many cars in the image?'', the attention map filters out the unrelated regions, which makes it easier for the model to infer the number of objects in an image.
\item For color (and more general attribute) questions, such as ``what is the color of the coat?'', the color of the specific object can be answered more effectively by focusing on the object of interest.
\item For object questions, such as ``what is sitting on top of the desk?'', the attention map can filter out less relevant regions such as background and infer better locations to look for objects according to their spatial relationship.
\item For location questions, such as ``where is the car in the image?'', the attention map is crucial for generating the correct answers because it evidently describes where the object is in the image.
\end{itemize}

\subsection{Question Understanding}\label{subsec: query_pro}
Question understanding is crucial for visual question answering.
The semantic meaning of questions not only provides the most important clue for answer generation, but also determines the CCKs to generate attention maps.

\begin{figure}
\begin{center}
\includegraphics[width=3.5in]{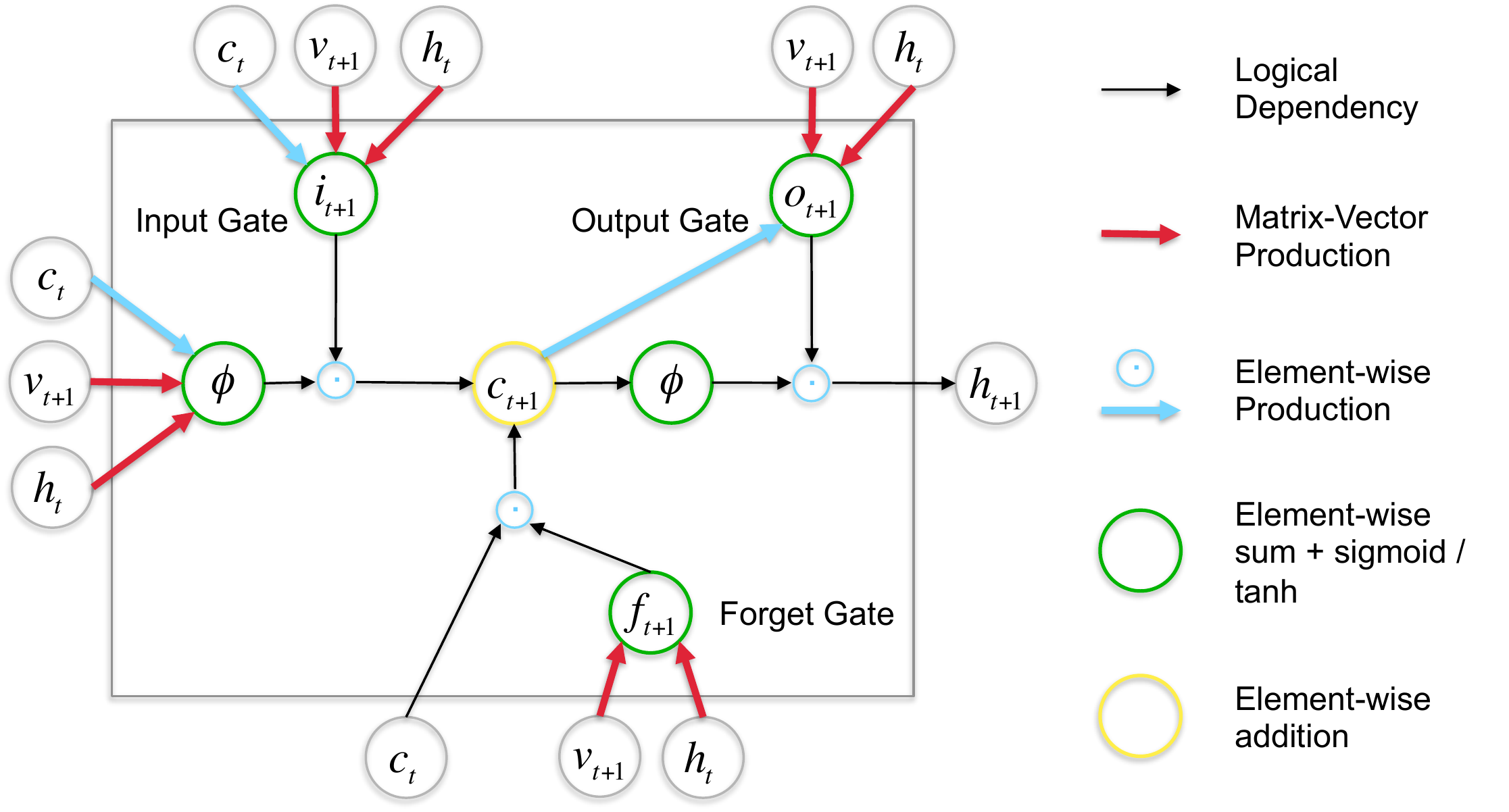}
\end{center}
   \caption{The structure of LSTM \cite{hochreiter1997long} for query processing}
\label{fig: lstm}
\end{figure}

Recently, LSTM model has shown good performances in language understanding \cite{hochreiter1997long}.
We employ an LSTM model to generate a dense question embedding to characterize the semantic meaning of questions.
A question $\bm{q}$ is first tokenized into word sequence $\{\bm{v}_t\}$.
We convert all the upper case characters to lower case characters, and remove all the punctuations.
The words that appear in training set but are absent in test set are replaced with a special symbol $\#OOV\#$.
Besides, $\#B\#$ and $\#E\#$ special symbols are added to the head and end of the sequence.
According to a {\em question dictionary}, each word is represented as a dense word embedding vector, which is learned in an end-to-end way.
An LSTM is applied to the word embedding sequence to generate a state $\bm{h}_t$ from each vector $\bm{v}_t$, using memory gate $\bm{c}_t$ and forget gate $\bm{f}_t$, which is illustrated in Eq.~\ref{equ: lstm equs}.
\begin{align}\label{equ: lstm equs}
\begin{split}
\bm{i}_t &= \sigma(\mathbf{W}_{vi}\bm{v}_t+\mathbf{W}_{hi}\bm{h}_{t-1}+\bm{b}_i)\\
\bm{f}_t &= \sigma(\mathbf{W}_{vf}\bm{v}_t+\mathbf{W}_{hf}\bm{h}_{t-1}+\bm{b}_f)\\
\bm{o}_t &= \sigma(\mathbf{W}_{vo}\bm{v}_t+\mathbf{W}_{ho}\bm{h}_{t-1}+\bm{b}_o)\\
\bm{g}_t &= \phi(\mathbf{W}_{vg}\bm{v}_t+\mathbf{W}_{hg}\bm{h}_{t-1}+\bm{b}_g)\\
\bm{c}_t &= \bm{f}_t\odot\bm{c}_{t-1}+\bm{i}_t\odot\bm{g}_t\\
\bm{h}_t &= \bm{o}_t\odot\phi(\bm{c}_t)\\
\end{split}
\end{align}
where $\phi$ is the hyperbolic tangent function and $\odot$ represents the element-wise production between two vectors. The LSTM's structure is shown in Fig.~\ref{fig: lstm}.
The semantic information $\bm{s}$ of the input question $\bm{q}$ is obtained by averaging the LSTM states $\{\bm{h}_t\}$ over all time steps .

\subsection{Image Feature Extraction}\label{subsec: img_ext}
The visual information in each image is represented as an $N \times N \times D$ image feature map.
The feature map is generated by dividing an image into an $N\times N$ grid, and extracting a $D$-dimensional feature vector $f$ in each cell of the grid.
The VGG-19~\cite{simonyan2014very} deep convolutional neural network extracts a $D$-dimensional feature vector for each window. The $D$-dimensional feature vector for each cell is the average of all the 10 $D$-dimensional feature vectors.
The final $N \times N \times D$ image feature map is the concatenation of $N\times N\times D$-dimensional feature vectors.

It is also possible to exploit a fully convolutional neural network architecture~\cite{long2014fully} to extract image feature maps more efficiently. We employ the segmentation model~\cite{chen2014semantic} pretrained on PASCAL 2007 segmentation dataset and find it leads to slightly better performance.

\subsection{Answer Generation}\label{subsec: ans_gen}
The answer generation part is a multi-class classifier based on the original image feature map, the dense question embedding, and the attention weighted feature map.

We employ the attention map to spatially weight the image feature map $\mathbf{I}$.
The weighted image feature map focuses on the objects asked in the question.
The spatial weighting is achieved by the element-wise production between each channel of the image feature map and the attention map.

\begin{equation}\label{equ: refine map}
\mathbf{I}^\prime_i = \mathbf{I}_i\odot \bm{m}
\end{equation}
where $\odot$ represents element-wise production.
$\mathbf{I}^\prime_i$ and $\mathbf{I}_i$ represent the $i$-th channel of attention weighted feature map $\mathbf{I}^\prime$ and original image feature map $\mathbf{I}$, respectively.
The attention weighted feature map lowers the weights of the regions that are irrelevant to the meaning of question.
To avoid overfitting, we apply an 1$\times$1 convolution on the attention weighted feature map to reduce the number of channels, resulting in a reduced feature map $\mathbf{I}_r$.
The question's semantic information $\bm{s}$, the image feature map $\mathbf{I}$ and the reduced feature map $\mathbf{I}_r$ are then fused by a nonlinear projection.
\begin{equation}
\bm{h} = g(\mathbf{W}_{ih}\mathbf{I}+\mathbf{W}_{rh}\mathbf{I}_r+\mathbf{W}_{sh}\bm{s}+\bm{b}_h)
\end{equation}
where $\bm{h}$ denotes the final projected feature, and $g(.)$ is the element-wise scaled hyperbolic tangent function: $g(x) = 1.7159\cdot\tanh(\frac{2}{3}x)$ \cite{lecun2012efficient}.
This function leads the gradients into the most non-linear range of value and enables a higher training speed.

A multi-class classifier with softmax activation, which is trained on the final projected features, predicts the index of an answer word specified in an answer dictionary.
The answer generated by ABC-CNN is the word with the maximum probability.
\begin{equation}\label{equ: ans generate}
a^* = \arg\max_{a\in\mathcal{V}_a} \bm{p}_{a} \text{\ \ \ s.t.\ \ \ }
\bm{p}_a = g(\mathbf{W}_{ha}\bm{h}+\bm{b}_{a})
\end{equation}
Notice that we do not share the word dictionary for questions and answers, i.e., one word can have different indices in the question dictionary and answer dictionary.

\subsection{Training and Testing}\label{subsec: train_test}
Our whole framework is trained in an end-to-end way with stochastic gradient descent and \emph{adadelta}~\cite{zeiler2012adadelta} algorithm.
Each batch of the stochastic gradient descent randomly samples 64 image question pairs independently, and back propagation is applied to learn all the weights of the ABC-CNN architecture.
We randomly adjust the initialization weights of all the layers to ensure that each dimension of the activations in all layers has zero mean and one standard variation.
The initial learning rate is set to be 0.1.
In our experiments, the weights in image feature extraction part are fixed to allow faster training speed, although it is possible to train all the weights in ABC-CNN in an end-to-end way.

During the testing stage, an image feature map is extracted for each image.
Given a question, we can produce its dense question embedding, and utilize the question embedding to configure the CCK to generate the attention map.
The multi-class classifier generates the answer using the original feature map, the question embedding, and the attention weighted feature map.

\section{Experiments}
We evaluate our model on Toronto COCO-QA \cite{torontoQa}, DAQUAR \cite{malinowski2014nips} and VQA datasets \cite{antol2015vqa}.
We evaluate our method on the QA pairs with single word answers, which accounts for (100\%, 85\%, 90\%) of Toronto-QA, VQA, DAQUAR datasets, respectively.
It is also consistent with the evaluation in ~\cite{torontoQa}.
Besides, our framework can be easily extended to generate answers in full sentences by using an RNN decoder in the answer generation part.

\subsection{Implementation Details}\label{sssec: exp_detail}
In experiments, we first choose the resolution of both the image feature map and the attention map to be 3$\times $3, which is called ``ATT'' model.
Each image cell generates a 4096-dimensional image feature vector using a pre-trained \emph{VGG} network~\cite{chatfield2014return}, and we extend each feature vector with the HSV histogram of the cell, resulting in a 4276-dimensional image feature vector for each cell.
The image feature vectors from all the image cells constitute an image feature map with dimension 4276$\times$3$\times$3.
To avoid overfitting, we reduce the dimension of the feature map to 256$\times$3$\times$3 with an 1$\times$1 convolution.
The dimension of the dense question embedding is 256.
We also try a second model called ``ATT-SEG'', which employs a fully convolutional neural network~\cite{chen2014semantic} pretrained on PASCAL 2007 segmentation dataset to generate 16$\times$16$\times$1024 feature maps, and concatenates them with HSV histograms in each cell as image feature maps.
In the end, we combine the VGG features, HSV features and segmentation features together, obtaining a model called ``ATT-VGG-SEG''.
It takes around 24 hours to train the network ATT on Toronto COCO-QA dataset with four K40 Nvidia GPUs.
The system can generate an answer at 9.89 ms per question on a single K40 GPU.

\subsection{Datasets}\label{sssec: exp_data}
We evaluate our models on three datasets: DAQUAR \cite{malinowski2014nips}, Toronto COCO-QA \cite{torontoQa} and VQA \cite{antol2015vqa}.

DAQUAR dataset has two versions: the full dataset (DQ-full) and the reduced one (DQ-reduced).
DQ-reduced has question answer pairs of 37 object classes, which is a subset of  DQ-full dataset that has 894 object classes.
Both versions use the indoor scenes images from NYU-Depth V2 dataset \cite{silberman2012indoor}.
The DQ-full dataset contains 795 training images with 6794 QA pairs, and 654 test images with 5674 QA pairs.
The DQ-reduced dataset contains 781 training images with 3825 QA pairs and 25 test images with 286  QA pairs.
We only train and test DAQUAR dataset on QA pairs with single word answers, which is consistent with the evaluation in \cite{torontoQa}.
Such QA pairs constitute (90.6\%, 89.5\%) and (98.7\%, 97.6\%) in the training and test sets for DQ-full and DQ-reduced datasets, respectively.

Toronto COCO-QA dataset uses images from Microsoft COCO dataset \cite{lin2014microsoft} (MS-COCO).
Its QA pairs only contain single word answers. Its basic statistics is summarized in Table \ref{tab: toronto_qa}.

\begin{figure*}
\begin{center}
\includegraphics[width=7.0in]{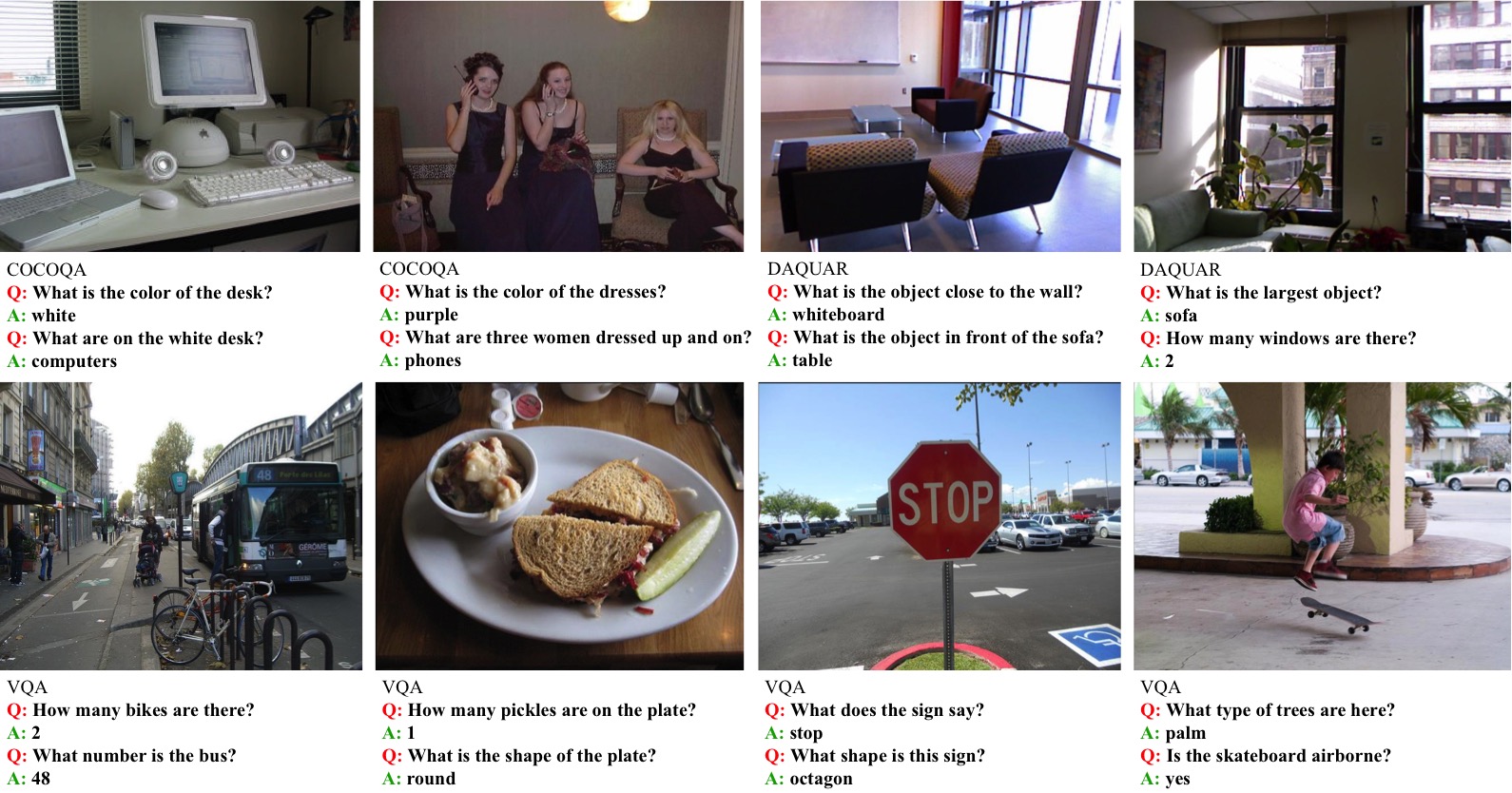}
\end{center}
   \caption{Example images and image-related QA pairs in Toronto COCO-QA  dataset \cite{torontoQa}, DAQUAR dataset \cite{malinowski2014nips} and VQA dataset \cite{antol2015vqa}. For VQA dataset, every question has 10 candidate answers. We show the answer with most votes for each question.}
\label{fig: dataset demo}
\end{figure*}

VQA dataset~\cite{antol2015vqa} is a recently collected dataset which is also built with images in MS-COCO dataset.
We evaluate the proposed model on VQA Real Image (Open-Ended) task in VQA dataset.
It contains 82783 training images, 40504 validation images, and 81434 testing images.
Each image in MS-COCO dataset is annotated with 3 questions, and each question has 10 candidate answers.
The total number of QA pairs for training, testing, and validation is 248349, 121512, 244302, respectively.
We only evaluate our method on the single-word answer QA pairs in VQA dataset, which constitute 86.88\% of the total QA pairs in this dataset.
Some examples from the three datasets are shown in Fig.~\ref{fig: dataset demo}.
\begin{table}
\begin{center}
\begin{tabular}{lcccc}
\hline
\textbf{Category} & \textbf{Train} & \textbf{\%} & \textbf{Test} & \textbf{\%} \\
\hline
Object & 54992 & 69.84 & 27206 & 69.85 \\
Number & 5885 & 7.47 & 2755 & 7.07\\
Color & 13059 & 16.59 & 6509 & 16.71 \\
Location & 4800 & 6.10 & 2478 & 6.36 \\
\hline
Total & 78736 & 100.00 & 38948 & 100.00\\
\hline
\end{tabular}
\end{center}
\caption{Toronto COCO-QA question type break-down~\cite{torontoQa}. }
\label{tab: toronto_qa}
\end{table}

\subsection{Evaluation Metrics}\label{sssec: exp_metric}
As in \cite{torontoQa}\cite{malinowski2014nips}, we evaluate the performance of the VQA models with ``answer accuracy'' (ACC.) and ``Wu-Palmer similarity measure Set'' (WUPS) score~\cite{wu1994verbs}\cite{malinowski2014nips}.
The answer accuracy computes the percentage of the generated answers that exactly match the ground truth answers.
The WUPS score is derived from the Wu-Palmer (WUP) similarity  \cite{wu1994verbs}, whose value is in the range of $[0, 1]$.
WUP similarity measures the similarity of two words based on the depth of their lowest common ancestor in the taxonomy tree~\cite{wu1994verbs}.
The WUPS score with threshold is the average of the down-weighted WUPS score for all the generated answers and ground truth.
If WUPS score of two words $s_{wups}$ is below a threshold, their down-weighted WUPS score is $0.1 s_{wups}$. Otherwise, its down-weighted WUPS is $s_{wups}$.
We use WUPS scores with thresholds 0.0 and 0.9 in our experiments, which are the same as \cite{malinowski2014nips}.



\subsection{Baseline Methods}\label{sssec: exp_base}
We compare the proposed method with different benchmark methods used in \cite{malinowski2014nips}\cite{torontoQa}\cite{antol2015vqa}\cite{malinowski2015ask}.
All the baseline models are listed below:
\begin{itemize}
\item \textbf{VIS+LSTM (VL)}: It is the framework proposed in \cite{torontoQa}, with a CNN extracting image features followed by a dimension reduction layer.
The image features are then inserted into the head position of the question word embedding sequences as inputs for question LSTM.
\item \textbf{2-VIS+BLSTM (2VB)}: The image features are inserted at the head and the tail of question word embedding sequences.
Besides, the question LSTM in \cite{torontoQa} is set to go in both forward and backward directions.
\item \textbf{IMG+BOW (IB)}: Ren et al.~\cite{torontoQa} use Bag-of-Words features to generate the dense question embedding.
\item \textbf{IMG}: Only the image features are used for answering the questions. It is called a ``deaf'' model.
\item \textbf{LSTM}: The answers are generated only using the dense question embedding from  LSTM. It is called a ``blind'' model.
\item \textbf{ENSEMBLE}: Ren et al. in \cite{torontoQa} evaluated the fusion model by using an ensemble of all the above methods.
\item \textbf{Q+I}: In \cite{antol2015vqa}, the question answering is achieved by training a multi-class classifier using both the dense question embeddings and image features.
\item \textbf{Q+I+C}: Compared to Q+I model in \cite{antol2015vqa}, the Q+I+C model \cite{antol2015vqa} adopts the dense embeddings of labeled image captions as an additional input.
\item \textbf{ASK}: In \cite{malinowski2015ask}, the answers are generated by linearly combining CNN features and question embeddings in an LSTM decoder.
\end{itemize}

\subsection{Results and Analysis}\label{sssec: exp_analysis}
Tables \ref{tab: toronto_qa_res}, \ref{tab: dqr_res}, and \ref{tab: dq_res} summarize the performance of different models on Toronto COCO-QA, DQ-reduced and DQ-full datasets, respectively.
Table~\ref{tab: toronto_qa_res} breaks down the performance of different methods in each category on Toronto COCO-QA dataset.

\begin{table*}
\begin{center}
\begin{small}
\begin{tabular}{l | cccc | ccc}
\hline
\textbf{Model} & \textbf{Object} & \textbf{Number} & \textbf{Color} & \textbf{Location} & \textbf{ACC.} & \textbf{WUPS 0.9} & \textbf{WUPS 0.0}\\
\hline
LSTM \cite{torontoQa}  & 0.3587 & 0.4534 & 0.3626 & 0.3842 & 0.3676 & 0.4758 & 0.8234 \\
IMG \cite{torontoQa} & 0.4073 & 0.2926 & 0.4268 & 0.4419 & 0.4302 & 0.5864 & 0.8585\\
IB \cite{torontoQa} & 0.5866 & 0.4410 & \textbf{0.5196} & 0.4939 & 0.5592 & 0.6678 & 0.8899 \\
VL \cite{torontoQa} & 0.5653 & 0.4610 & 0.4587 & 0.4552 & 0.5331 & 0.6391 & 0.8825 \\
2VB \cite{torontoQa} & 0.5817 & 0.4479 & 0.4953 & 0.4734 & 0.5509 & 0.6534 & 0.8864 \\
ENSEMBLE \cite{torontoQa} & 0.6108 & 0.4766 & 0.5148 & 0.5028 & 0.5784 & 0.6790 & 0.8952 \\
\hline
NO-ATT & 0.5882 & 0.4319 & 0.4168 & 0.4762 & 0.5414 & 0.6483 & 0.8855\\
ATT & 0.6217 & \textbf{0.4799} & 0.4727 & 0.5194 & 0.5803 & 0.6814 & 0.8966 \\
ATT-SEG & 0.6238 & 0.4617 & 0.4694 & 0.5278 & 0.5804 & 0.6833 & 0.8979\\
ATT-VGG-SEG & \textbf{0.6246} & 0.4570 & 0.4681 & \textbf{0.5367} & \textbf{0.5810} & \textbf{0.6844} & \textbf{0.8985}\\
\hline
\end{tabular}
\end{small}
\end{center}
\caption{Results on Toronto COCO-QA dataset \cite{torontoQa} in four different kinds of questions and overall accuracy (ACC.), WUPS score at threshold of 0.9 (WUPS 0.9) and 0.0 (WUPS 0.0)}
\label{tab: toronto_qa_res}
\end{table*}

In Table \ref{tab: toronto_qa_res}, ABC-CNN using only ATT model surpasses all the baseline models.
It also outperforms the ENSEMBLE model  by 0.3\% in term of answer accuracy, although we only employ a single model.
The ABC-CNN outperforms the baseline methods in ``object'', ``number'' and ``location'' categories, because question-guided attention exploits semantics of questions and the contextual information in images to answer the questions.
Its accuracy is slightly lower than IB and ENSEMBLE models in the ``color'' category.
We also find the performance of the fully convolutional model ATT-SEG is slightly better than ATT, while extracting feature maps with fully convolutional neural networks is much faster.
Combination of the features in ATT and ATT-SEG together (ATT-VGG-SEG) results in the best performance.
In particular, adding fully convolutional model helps correctly answer the location questions.
We also try to remove the attention in ABC-CNN (NO-ATT) as an ablative experiment, and it results in 1.34\% , 0.85\%, and 0.35\% loss in accuracy, WUPS 0.9 and WUPS 0.0 scores, respectively.

In Table \ref{tab: dqr_res}, we compare ABC-CNN model with the baseline models on DQ-reduced dataset. Its performance is higher than all the single models on all the metrics. It is only 0.53\% lower than the ENSEMBLE model on WUPS 0.9 measure.

On DQ-full and VQA datasets, ABC-CNN outperforms state-of-the-art methods on both datasets in Table. \ref{tab: dq_res} and \ref{tab: vqa_res}.
On DQ-full dataset, the ABC-CNN model is the same as the models on Toronto COCO-QA and DQ-reduced dataset.
On VQA dataset, to make a fair evaluation, we employ the same answer dictionary that contains the 1000 most frequent answers (ATT 1000) as \cite{antol2015vqa}.
We also evaluate the ABC-CNN model using the answer dictionary that  contains all the answers (ATT Full).
\begin{table}
\begin{center}
\begin{small}
\begin{tabular}{lccc}
\hline
\textbf{Model} & \textbf{ACC.} & \textbf{WUPS 0.9} & \textbf{WUPS 0.0} \\
\hline
LSTM \cite{torontoQa}& 0.3273 & 0.4350 & 0.8162 \\
IB \cite{torontoQa}& 0.3417 & 0.4499 & 0.8148 \\
VL \cite{torontoQa}& 0.3441 & 0.4605 & 0.8223 \\
2VB \cite{torontoQa}& 0.3578 & 0.4683 & 0.8215\\
ENSEMBLE \cite{torontoQa} & 0.3694 & \textbf{0.4815} & 0.8268 \\
\hline
NO-ATT & 0.3931 & 0.4445 & 0.8230 \\
ATT & \textbf{0.4276} & 0.4762 & \textbf{0.8304}\\
\hline
HUMAN & 0.6027 & 0.6104 & 0.7896\\
\hline
\end{tabular}
\end{small}
\end{center}
\caption{Results on DAQUAR-reduced dataset \cite{malinowski2014nips}}
\label{tab: dqr_res}
\end{table}
\begin{table}
\begin{center}
\begin{tabular}{lccc}
\hline
\textbf{Model} & \textbf{ACC.} & \textbf{WUPS 0.9} & \textbf{WUPS 0.0} \\
\hline
ASK \cite{malinowski2014nips} & 0.1943 & 0.2528 & 0.6200 \\
ATT & \textbf{0.2537} & \textbf{0.3135} & 0.6589 \\
\hline
HUMAN & 0.5020 & 0.5082 & 0.6727 \\
\hline
\end{tabular}
\end{center}
\caption{Results on DAQUAR-full dataset \cite{malinowski2014nips}}
\label{tab: dq_res}
\end{table}

\begin{figure*}
\begin{center}
\includegraphics[width=7.0in]{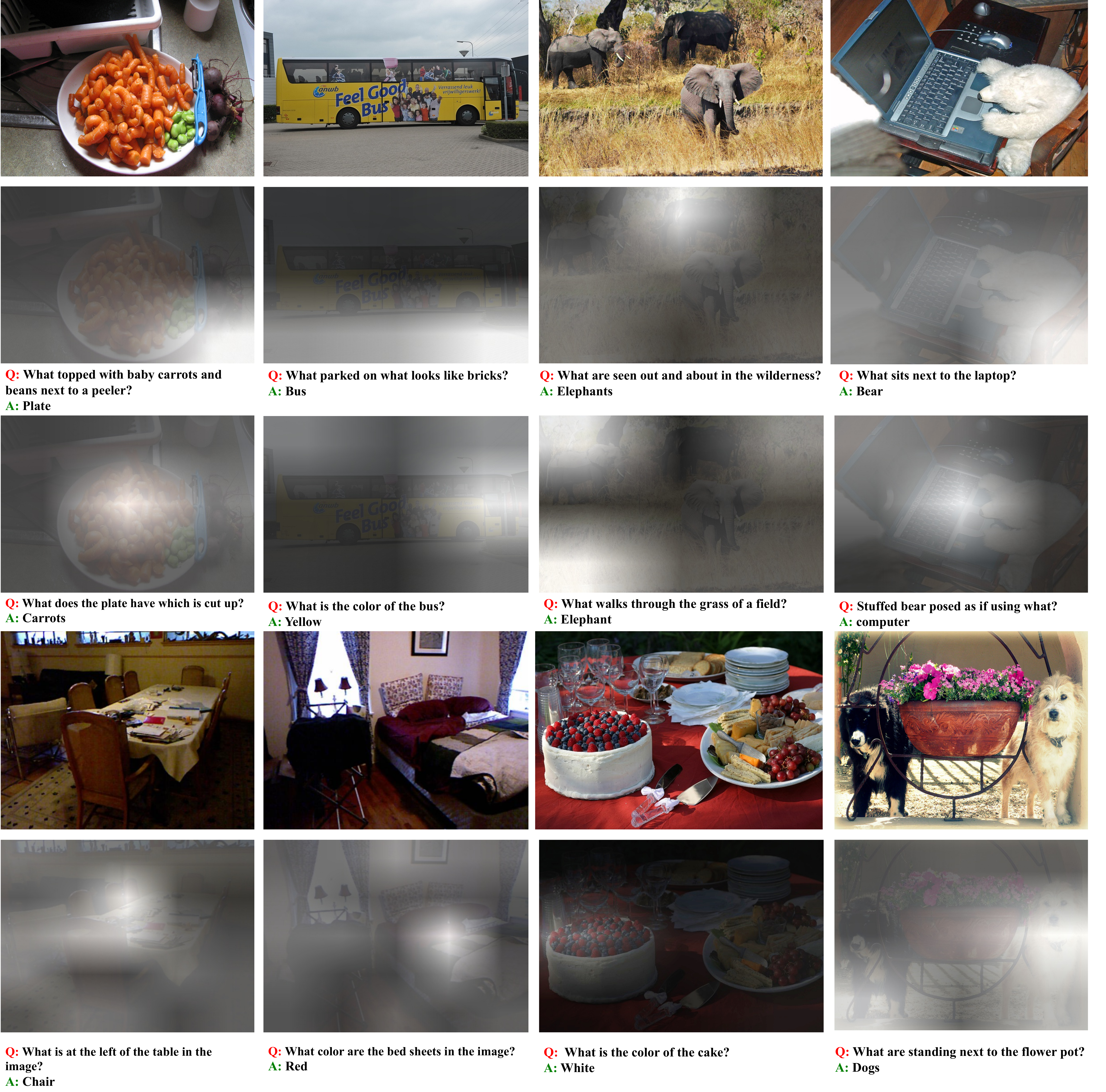}
\end{center}
   \caption{Selected images with image-related questions and question-guided attention maps generated by ABC-CNN in Toronto COCO-QA dataset \cite{torontoQa}. We find the proposed ABC-CNN model is capable of focusing its attention on different regions for different questions. The attention maps help filter out irrelevant information and model the spatial relationship of the objects in images.}
\label{fig: result demo}
\end{figure*}

Some of the generated question-guided attention maps and their corresponding images and questions  are shown  in Fig. \ref{fig: result demo}.
We can observe that the question-guided attention maps successfully capture different questions' intents with different attention regions.
With these attention maps, ABC-CNN is capable of generating more accurate answers by focusing its attention on important regions and filtering out irrelevant information.
Since the original feature map is also provided when predicting answers, ABC-CNN can answer the question without using the attention map if the object queried is the only object in the image (e.g., the QA pair in row 2, column 2 of Fig.~\ref{fig: result demo}). 
In this case, the attention map may not focus on the object queried.


\begin{table}
\begin{center}
\begin{tabular}{lcccc}
\hline
\textbf{Model} & Q+I~\cite{antol2015vqa} & Q+I+C~\cite{antol2015vqa} & ATT 1000 & ATT Full \\
\hline
\textbf{ACC.} & 0.2678 & 0.2939 & \textbf{0.4838} & 0.4651 \\
\hline
\end{tabular}
\end{center}
\caption{Results on VQA dataset~\cite{antol2015vqa}}
\label{tab: vqa_res}
\end{table}

\section{Conclusion}
In this paper, we propose a unified attention based configurable convolutional neural network (ABC-CNN) framework for VQA problem.
Our model unifies the visual feature extraction and semantic question understanding via \emph{question-guided} attention map.
The attention map is generated by a configurable convolution network that is adaptively determined by the meaning of questions.
ABC-CNN significantly improves both visual question answering performance and the understanding of the integration of question semantics and image contents.
Our model outperforms state-of-the-art methods on Toronto COCO-QA, DAQUAR and VQA datasets.
The visualization demonstrates that ABC-CNN produces attention maps that are highly related to the semantics of the image-related questions.

{\small
\bibliographystyle{ieee}
\bibliography{egbib}
}

\end{document}